\pdfoutput=1

\documentclass[11pt]{article}

\usepackage[]{acl}

\usepackage{times}
\usepackage{latexsym}
\usepackage{graphicx}
\usepackage{enumitem}
\usepackage{float}
\usepackage{amsmath}

\usepackage[T1]{fontenc}

\usepackage[utf8]{inputenc}

\usepackage{microtype}

\title{ Analyzing sports commentary in order to automatically recognize events and extract insights}

\author{Yanis Miraoui \\
        ETH Zürich  \\
        \texttt{ymiraoui@student.ethz.ch} \\
        January, 2022
        }

\begin{document}
\maketitle
\begin{abstract}
In this paper, we carefully investigate how we can use multiple different Natural Language Processing techniques and methods in order to automatically recognize the main actions in sports events. We aim to extract insights by analyzing live sport commentaries from different sources and by classifying these major actions into different categories. We also study if sentiment analysis could help detect these main actions. 
\end{abstract}

\vspace{3 pt}

\section{Introduction}

This research report aims to study how to recognize and analyze the main actions in sports events. \\
Indeed, detecting important actions and extracting insights of sports events has become of great interest in the recent years. Extracting these insightful moments allows for example sports teams to analyze their performance meticulously, media broadcasters to repurpose and redistribute their video content easily and fans to get more engaged. \\
On the other hand, during most sports events, media broadcasters provide their audience with very explanatory live commentaries through different means (TV, radio, newspaper, website, ...). These live commentaries have been particularly well developed for Football (Soccer) : Almost every sports newspaper or \emph{live score website}\footnote{ E.g., \href{https://www.livescore.com/}{livescore.com}, \href{https://www.flashscore.com/}{flashscore.com}, \href{https://www.sofascore.com/}{sofascore.com}, …} has now a section dedicated to live football commentary. \\
Therefore, in this study, we intend to use these large amount of audio and textual live commentaries to detect the main actions in sports events. This would allow us to automatically extract the most insightful moments in sports games. \\
The first step of our study is to build our dataset. In order to use audio live commentary, we have to transcribe the sports commentary audio into exploitable textual data. We would then be able to perform a thorough analysis on textual content. Moreover, we also make use of textual live commentaries from live score websites. Equipped with this textual data, we apply the main preprocessing techniques learned in this course. We then train different classification models able to detect and classify actions during a sport event. Finally, we evaluate the metrics of these models on different types of live sports commentaries and determine how this model could have a real impact on the performance of an already existing event recognition AI. We also explore if sentiment analysis could help recognize important actions during some sports events.

\begin{figure}[h]
    \hspace{-10pt}\includegraphics[scale=0.29]{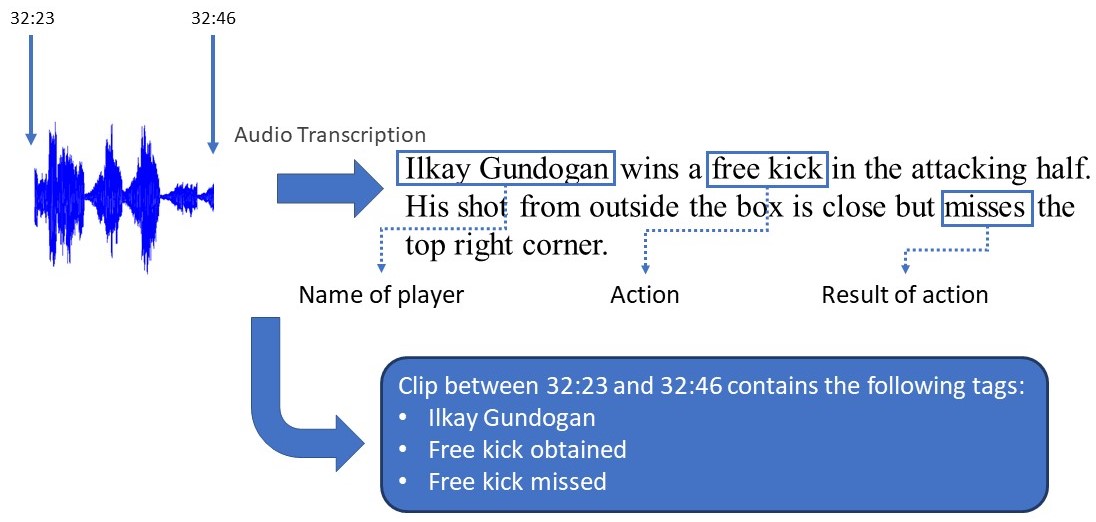}
    \caption{Schema showing how an ideal simple example of audio live sports commentary would be analyzed and processed.}
    \label{fig:example}
\end{figure}

\section{Literature review and related work}

Only a very few works and research have been done on this subject. This is not very surprising as this topic emerged only recently and remains very niche and specialized.  \\
However, one research paper in particular, aims to study these live sports commentaries with a similar method \citep{AMS}. Indeed, this study mainly focus on trying to understand the content of live sports commentaries by detecting and classifying relevant events in football games. This paper mainly used newspapers live commentaries that they manually annotated. Moreover, they used a SVM model to classify the events in different categories and sub-categories. As this research paper has a very similar intended objective, we will make use of its methods and its model as a baseline for our study.\\
It is also possible to find several research papers that aim to study the automatic generation of live sports commentaries for video games \citep{MZDK} or media broadcasters \citep{ANR}. Some research papers study as well how to automatically generate sports news article by summarizing the main actions of sports events using pre-existing datasets \citep{KHC} and \citep{ZYX} or by analyzing Tweets \citep{JJC}. These studies remain relevant regarding the semantic analysis they perform. Nevertheless, the methods and techniques used do not fit well with our intended purpose. \\
On the other hand, we can notice that there has been an important interest from the NLP community for event recognition and classification through the organization of several hackathons:
\vspace{-9pt}
\begin{itemize}[noitemsep]
    \item Hackathon organized by Google in Paris in 2016, \href{http://hackatal.github.io/2016/}{http://hackatal.github.io/2016/}
    \item Hackathon organized by the United States Soccer Federation in 2015, \href{https://www.ussoccer.com/official-us-soccer-hackathon/}{http://www.ussoccer.com/official-us-soccer-hackathon/}
\end{itemize}

\section{Background}

This study is carried out in link with a company (\emph{Egoli Media}\footnote{\emph{Egoli Media} is a proprietary AI-enabled video annotation technology which enables real-time personalized delivery of video content. This AI aims to annotate videos by detecting the main events and athletes in sports content. For more information: \href{https://www.egolimedia.com/}{egolimedia.com}.}) which kindly provided us an important amount of data. \\ 
Indeed, the need of such a study arises as the company is currently trying to improve its event recognition pipeline. One of the ways to reach that goal would be to optimize the computer vision AI inspecting video content. However, this improvement is very limited. On the other hand, we also have access to the English sports commentary of most of the sports events. Within the company, no work has already been done regarding how to make use of this large amount of sports commentaries to improve the event recognition.  \\
Moreover, from a wider viewpoint, many different live scores websites have been recently developing and using these live sports commentaries to provide their audience with a more detailed and accurate description of the main actions of a sports event. These live commentaries aim to truly engage fans by giving them more absorbing insights.

\section{Data sources}

\subsection{Audio dataset}
We obtained from the company \emph{Egoli Media} a rich audio dataset. This dataset is extracted from real live sports commentary of the 2021 Paralympic games and of a few English Premier League games. To transcribe the audio into text, we used the Speech-to-Text API developed by Google \citep{Google}. This tool aims to convert speech into text by using Google’s AI technologies. However, when inspecting the transcription, the dataset seems to be very noisy. It would be very laborious to exploit it, for example:
\begin{itemize}
    \item \emph{“Like that shot that got put in on him by the three americans as a work in a dormant volcano cause he's beginning to bubble”}
    \item \emph{“Babies and kids the and some of the united states of america so do about it goes as you could be easily”}
    \item \emph{“The first minute so gym roberts surveys the scene the usa team happy it's a cluster the key and make things awkward for gb but robots finds is running part of our and fits”}
\end{itemize}

We notice that this transcription looks very inaccurate and unreliable. Building a high-performing event recognition and classification model with this data seems very difficult.

\subsection{Textual dataset}

Therefore, we decide to mostly use other sources of data for building and training a model that would perform well. \\
Firstly, we build a complete dataset of clean text commentaries by scraping live scores websites such as \href{https://www.bbc.com/sport}{bbc.com}, \href{https://espn.com/}{espn.com} and \href{https://onefootball.com/}{onefootball.com}. These live commentaries are only extracted from football events as they are the most enriched textual commentaries. These commentaries have the advantage to be as well accompanied with a timeline of events that would represent the labels for our training process. The meaning and definition of these labels can be found in \href{sec:appendix}{Appendix A}. \\ 

\setlength{\tabcolsep}{0.5em} 
\renewcommand{\arraystretch}{1.3}
\begin{table}[H]
\begin{tabular}{lc}
\hline
League & Number of games  \\ \hline

Serie A (Italy) & 2,152\\
Ligue 1 (France) &  2,076\\
La Liga (Spain) &  1,939\\
Bundesliga (Germany) &  1,608\\
 Premier League (UK) &  1,299\\ \hline
TOTAL &  9074 \\
\hline
\end{tabular}
\caption{\label{tab:leaguesandgames} Distribution  of the number of games of our dataset in terms of the different football leagues.}
\end{table}
\vspace{-10pt}
Moreover, we can also consider building the training dataset by scraping the automatic subtitles add-on of YouTube. Indeed, it is also possible to obtain the textual live commentaries from various sports events through the YouTube API. Nonetheless, this method remains laborious as it would imply to manually label a large amount of sentences for the training of our model. Therefore, we only use this method to test and evaluate the accuracy of our model on a different dataset.

\section{Modeling techniques explored}

\subsection{Cleaning and preprocessing of the textual data}
The dataset scraped from the live score websites contains $\approx$ 941,000 sentences labeled with one of the 12 categories (Attempt, Corner, Foul, Yellow Card, …). These textual commentaries are extracted from 9074 real games that occurred since 2011 in the five biggest European football leagues. An essential step in the preprocessing of this dataset was to check that the dataset was balanced enough to train a classification model on:
\vspace{-15pt}
\begin{figure}[H]
    \hspace{-20pt}\includegraphics[scale=0.6]{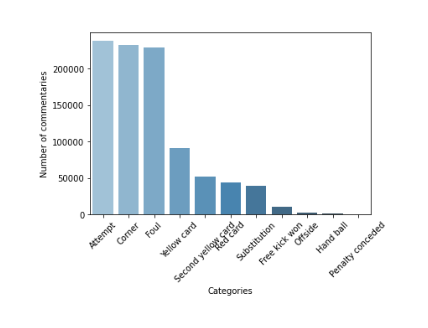}
    \vspace{-35pt}
    \caption{Bar plot showing the distribution of the different categories of the dataset of live textual commentaries.}
    \label{fig:catplot}
\end{figure}
\vspace{-5pt}

As we can notice on \href{fig:catplot}{Figure 2}, even if only a few commentaries belong to the categories “Offside”, “Hand ball” and “Penalty conceded”, there is not one category that holds more than 50\% of the commentaries. However, we could still consider re-balancing the dataset using oversampling \citep{SMOTE}. \\
Furthermore, we apply some simple textual cleaning techniques: 
\vspace{-3pt}
\begin{itemize}[noitemsep]
    \setlength\itemsep{0.01em}
    \item Removing punctuation, special characters, leading, trailing and extra white spaces/tabs
    \item Stop-word removal
    \item Stemming
    \item Lemmatization
\end{itemize}

\setlength{\tabcolsep}{0.2em} 
\begin{table}[H]
\begin{tabular}{lccc}
\hline
Statistics & Avg. words & Avg. characters \\ \hline
Live commentaries & 12.05 & 73.459 \\ \hline
\end{tabular}
\caption{\label{tab:statsdata} Statistics of our scraped textual dataset.}
\end{table}

\subsection{Modeling using tf-Idf}

On one hand, for our word embedding, we build vectors from this cleaned textual data using the Term Frequency-Inverse Document Frequencies (tf-Idf). Using this method seems logical because in order to classify the textual live commentaries, it is essential to encode the frequencies of the most important term in each sentence with their relevancy \citep{tfidf}. This is also the method that gives us the best results in building an accurate classification model. Equipped with these embedded sentences with tf-Idf, we are now able to train, evaluate and compare different classification algorithms.

\subsection{Modeling using BERT}

On the other hand, we also consider using BERT Transformers to model our text classification task. Indeed, BERT is designed to help computers understand the meaning of ambiguous language in text by using the bidirectional surrounding text to establish context \citep{bert}. Therefore, BERT is a State of the Art language model that distinguishes itself from previous language models such as word2vec and GloVe, which are limited when interpreting context and polysemous words.
This is very pertinent to our task. To classify the live commentaries efficiently, we would want our model to take into account the surrounding context of each word. This is why we will explore and evaluate this modeling technique more thoroughly. 

\section{Experimental findings and evaluation}

\subsection{Baseline : SVM model}
The research paper \citep{AMS} studies how an SVM model could be trained in order to detect and classify the main events in newspapers commentaries. As this study represents our baseline, we try to reproduce its methods and techniques using our large live commentaries dataset.\\
Firstly, we randomly split the tf-Idf embedded vectors into a train and a test subset using ratios of 80\% / 20\% respectively. We decide to shuffle the data when splitting, as we want our model to be well-functioning on every type of game and for every commentary made at anytime during the game. \\
We then train our SVM classifier model and evaluate its metrics on the test set:
\setlength{\tabcolsep}{0.3em} 
\renewcommand{\arraystretch}{1.3}
\begin{table}[H]
\begin{tabular}{lcccc}
\hline
Metrics: & Accuracy & Precision & Recall & F1 Score  \\ \hline
SVM & 97.30\% & 88.64\% & 83.60\% & 0.85\\
\hline
\end{tabular}
\caption{\label{tab:svmmetrics} Evaluation metrics of the SVM classifier model.}
\end{table}
We also build the confusion matrix comparing the predictions made by the SVM model on the test set and the true labels:
\vspace{-10pt}
\begin{figure}[H]
    \hspace{-5pt}\includegraphics[scale=0.33]{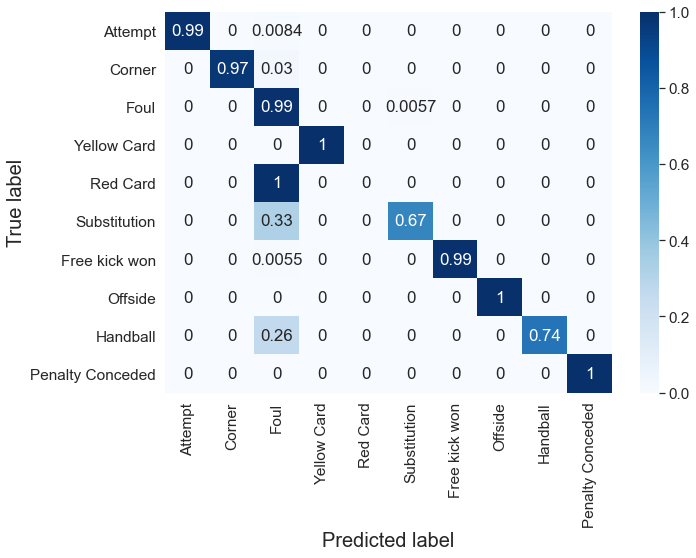}
    \vspace{-15pt}
    \caption{Confusion matrix of the evaluation of the SVM model.}
    \label{fig:svmconf}
\end{figure}
\vspace{-5pt}
At first sight, we can observe on \href{fig:svmconf}{Figure 3} that most of the errors concern commentaries that are mislabeled as \emph{"Foul"}. For example 26\% of the true labels \emph{"Handball"} are being mislabeled as \emph{"Foul"}. However, these errors can be caused by the fact that most handballs in football are considered to be fouls as well. Hence, the clear distinction between the categories \emph{"Foul"} and \emph{"Handball"} could be quite ambiguous in football.\\
On the other hand, we also notice that our SVM is still very powerful and can very accurately classify the main actions of various football games.
The baseline model built in the research paper achieved an F1 score of 0.71. In comparison, our SVM model attains an F1 score of 0.85 (\href{tab:svmmetrics}{Table 3}). This important improvement could be explained by the fact that we have trained our model on a much larger dataset.

\subsection{Other classification models using tf-Idf}
In order to extend our analysis, we build various other classification models using the same embedded vectors with tf-Idf \citep{classifiers}. We also use the same train/test splitting methods for training and evaluating the different classifiers. \\
To evaluate and compare these different classification models, we use \textbf{the accuracy} metric ($accuracy = \frac{\text{Number of correct predictions}}{\text{Total number of predictions}}$). For our task, we are willing to get the most right predictions for the commentary of each game. Therefore, we build and train the different classifications models. Firstly, we compare the accuracy of each model on the same type of live sports commentaries:

\vspace{-5pt}
\begin{figure}[H]
    \includegraphics[scale=0.5]{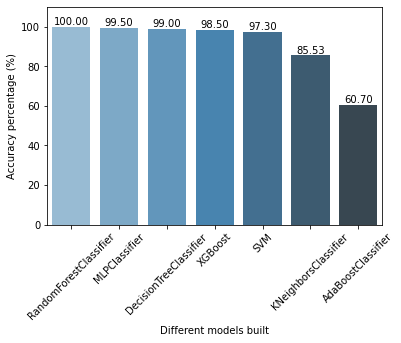}
    \vspace{-10pt}
    \caption{Histogram comparing the accuracy of different classification models.}
    \label{fig:histclas}
\end{figure}
\vspace{-5pt}
On the other hand, we want our classification model to be able to generalize well and to perform well on different types of live commentaries structures. For example, we would like our model to perform effectively on commentaries extracted from other sources like Youtube or \href{livescore.com}{livescore.com}.
Therefore, we also compare the accuracy of each model on new these types of live sports commentaries which have different semantic structures and syntactic expressions:
\vspace{-5pt}
\begin{figure}[H]
    \includegraphics[scale=0.48]{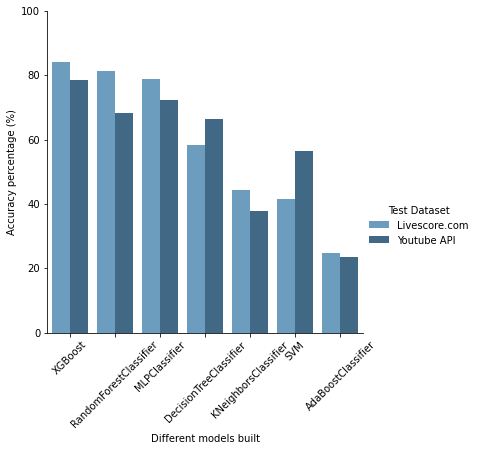}
    \vspace{-10pt}
    \caption{Histogram comparing the accuracy of different classification models on the new dataset scraped from \href{livescore.com}{livescore.com} and Youtube.}
    \label{fig:histclas_ext}
\end{figure}
\vspace{-5pt}
We deduce from this comparison that most classification models generalize relatively well to new types of sports commentaries (\href{fig:histclas_ext}{Figure 5}). Moreover, the XGBoost model may represent the best performing model for our task by accurately predicting the label of most of the events and from different types of live commentaries structures.

\subsection{BERT Classification model}
In addition, to model our classification task, we study how implementing a BERT classification model could improve our results. Indeed, we expect that the use of BERT Transformers would confer us a better accuracy as this modeling technique is taking into account the surrounding context of each word. Therefore, we train on \emph{Google Collab}\footnote{Google Collab provide us powerful CPU computation resources, essential for training a BERT classification model.} our model with our dataset using the existing transformer pretrained on a large corpus of texts (\emph{'bert-base-uncased'}). The dataset which we use to fine-tune the BERT model is similar to the one used previously. However, we do not proceed to the tf-Idf vectorizing step in this case.  \\
We evaluate our classification model on unseen live sports commentaries and obtain a very high accuracy of 99.8\%. This new model is also evaluated on new types of sentences structures using the dataset scraped from \href{livescore.com}{livescore.com} and Youtube. It attains a also very high accuracy of 92\% in average. 
\vspace{-5pt}
\begin{figure}[H]
    \includegraphics[scale=0.40]{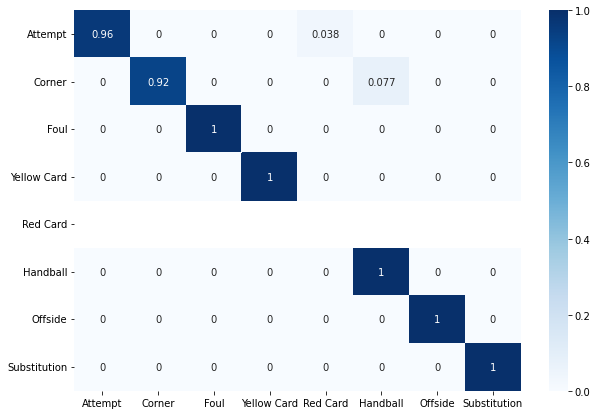}
    \vspace{-10pt}
    \caption{Confusion matrix of the evaluation of the fine-tuned BERT model.}
    \label{fig:bertconf}
\end{figure}
\vspace{-5pt}
Hence, modeling our classification task using a fine-tuned BERT transformer seems to be a very high-performing and powerful solution. This modeling technique, by accurately grasping the context of each word, seems to significantly improve our results. 

\subsection{Sentiment Analysis for Event Recognition}
Another less obvious alternative to detect the main actions during sports events is to perform a sentiment analysis of the live commentaries. In fact, we could think that scoring a goal (positive) does not convey the same emotions as conceding a red card (negative). This is why, we make use of \emph{'cardiffnlp/twitter-roberta-base-sentiment'} model to detect if some event types convey more positive or negative emotions when being commented. If this is the case, this would confirm our hypothesis that sentiment analysis of live commentaries could help detect and classify the main actions in many sports events.
\\ 
Therefore, we predict the sentiment conveyed by each sentence of a sample of our live sports commentaries ($\approx$10,000 sentences) using the pretrained model \emph{'cardiffnlp/twitter-roberta-base-sentiment'}. This model is a roBERTa-base model trained on $\approx$58M tweets and finetuned for sentiment analysis \citep{sentiment}. This pretrained model seems to be the best performing model for our task as there does not exist any pretrained sentiment analysis model specifically for sports. Moreover, this model is known for being very powerful for many general tasks\interfootnotelinepenalty=10000\footnote{Note that we could have built our own sentiment analysis model for sports. However, building this model would have been very time consuming as we would have to label manually an important amount of data.}. We obtain the following results by disposing of the {\color{gray}"Neutral"} labels and reporting only the predominant {\color{green}"Positive"} or {\color{red}"Negative"} label:

\setlength{\tabcolsep}{0.5em} 
\renewcommand{\arraystretch}{1.3}
\begin{table}[H]
\begin{tabular}{lc}
\hline
Event type & Sentiment\\ \hline
Attempt & \color{red}Negative (6.04\%)\\
Corner &  \color{green}Positive (2.59\%)\\
Foul &  \color{red}Negative (18.77\%)\\
Yellow card &  \color{red}Negative (74.01\%)\\
Second yellow card &  \color{red}Negative (38.32\%)\\
Red card &  \color{red}Negative (70\%)\\
Substitution &  \color{green}Positive (1.37\%)\\
Free kick won & \color{green}Positive (0.49\%)\\
Offside &  \color{green}Positive (2.97\%)\\
Handball &  \color{darkgray}None \\
Penalty conceded &  \color{red}Negative (100\%)\\
\hline
\end{tabular}
\caption{\label{tab:sentimentdistribution} Results of the sentiment analysis peformed on the different event types.}
\end{table}
\vspace{-10pt}

We first observe from \href{tab:sentimentdistribution}{Table 4} that the results obtained seem quite convincing. Indeed, a "Foul", a "Yellow card" or a "Red card" for example are events that convey a negative sentiment when commented. \\
On the other hand, we can also notice that most of the {\color{green}"Positive"} or {\color{red}"Negative"} labels have a very small percentage of appearance. This means that most live commentaries are predicted as being {\color{gray}"Neutral"}. If most of the live sports commentaries are predicted as being {\color{gray}"Neutral"}, the information gained by this sentiment analysis becomes very limited. Perhaps, building a specialized sentiment analysis model for sports could help improve our results and the relevancy of performing a sentiment analysis.

\section{Conclusion}
By carrying out this study, we analyzed how different Natural Language Processing techniques and methods could be used in order to automatically recognize events and extract insights during sports events. We mainly worked with live textual commentaries to build different classification models. In particular, we explored mainly two different techniques in order to embed the textual commentaries into numerical vectors. \\
Using the tf-Idf embeddings, we built an SVM model following the same method as in the baseline \citep{AMS}. This model obtained an F1 score of 0.85, outperforming the baseline model. Moreover, we built and compared different classification models before deciding on the most powerful one: the XGBoost model. \\
On the other hand, we studied two different approaches using the BERT Transformers. The first one aimed to label the live commentaries by using the bidirectional surrounding text to establish context. The fine-tuned classification model outperformed all the models previously built, becoming therefore the best model to solve our task with an accuracy of 92\% in average.
The second approach was to determine if sentiment analysis could help detect the main actions in sports events. However, this analysis is much more nuanced: the information gained by a sentiment analysis of the live commentaries could be very limited.

\section*{Acknowledgements}
We especially thank \emph{Egoli Media} for providing us with an important amount of data and for their feedback.

\bibliography{acl_latex}

\begin{thebibliography}{12}
\expandafter\ifx\csname natexlab\endcsname\relax\def\natexlab#1{#1}\fi

\bibitem[{Barbieri et~al.(2020)Barbieri, Camacho{-}Collados, Neves, and
  Anke}]{sentiment}
Francesco Barbieri, Jos{\'{e}} Camacho{-}Collados, Leonardo Neves, and
  Luis~Espinosa Anke. 2020.
\newblock \href {http://arxiv.org/abs/2010.12421} {Tweeteval: Unified benchmark
  and comparative evaluation for tweet classification}.
\newblock \emph{CoRR}, abs/2010.12421.

\bibitem[{Chawla et~al.(2002)Chawla, Bowyer, Hall, and Kegelmeyer}]{SMOTE}
N.~V. Chawla, K.~W. Bowyer, L.~O. Hall, and W.~P. Kegelmeyer. 2002.
\newblock \href {https://doi.org/10.1613/jair.953} {Smote: Synthetic minority
  over-sampling technique}.
\newblock \emph{Journal of Artificial Intelligence Research}, 16:321–357.

\bibitem[{Chiu et~al.(2018)Chiu, Sainath, Wu, Prabhavalkar, Nguyen, Chen,
  Kannan, Weiss, Rao, Gonina, Jaitly, Li, Chorowski, and Bacchiani}]{Google}
Chung-Cheng Chiu, Tara Sainath, Yonghui Wu, Rohit Prabhavalkar, Patrick Nguyen,
  Zhifeng Chen, Anjuli Kannan, Ron~J. Weiss, Kanishka Rao, Katya Gonina,
  Navdeep Jaitly, Bo~Li, Jan Chorowski, and Michiel Bacchiani. 2018.
\newblock \href {https://arxiv.org/pdf/1712.01769.pdf} {State-of-the-art speech
  recognition with sequence-to-sequence models}.
\newblock \emph{ICASSP 2018}.

\bibitem[{Devlin et~al.(2019)Devlin, Chang, Lee, and Toutanova}]{bert}
Jacob Devlin, Ming-Wei Chang, Kenton Lee, and Kristina Toutanova. 2019.
\newblock \href {http://arxiv.org/abs/1810.04805} {Bert: Pre-training of deep
  bidirectional transformers for language understanding}.
\newblock \emph{2019 Conference of the North American Chapter of the
  Association for Computational Linguistics}.

\bibitem[{Huang et~al.(2020)Huang, Li, and Chang}]{KHC}
Kuan-Hao Huang, Chen Li, and Kai-Wei Chang. 2020.
\newblock \href {https://aclanthology.org/2020.aacl-main.61.pdf} {Generating
  sports news from live commentary: A chinese dataset for sports game
  summarization}.
\newblock \emph{AACL}.

\bibitem[{Liu et~al.(2018)Liu, Sheng, Wei, and Yang}]{tfidf}
Cai-zhi Liu, Yan-xiu Sheng, Zhi-qiang Wei, and Yong-Quan Yang. 2018.
\newblock \href {https://doi.org/10.1109/IRCE.2018.8492945} {Research of text
  classification based on improved tf-idf algorithm}.
\newblock \emph{IEEE}, pages 218--222.

\bibitem[{Minard et~al.(2016)Minard, Speranza, Magnini, and Qwaider}]{AMS}
Anne-Lyse Minard, Manuela Speranza, Bernardo Magnini, and Mohammed~R.H.
  Qwaider. 2016.
\newblock \href {http://ceur-ws.org/Vol-1749/paper36.pdf} {Semantic
  interpretation of events in live soccer commentaries}.
\newblock \emph{Proceedings of the Third Italian Conference on Computational
  Linguistics}.

\bibitem[{Nichols et~al.(2012)Nichols, Mahmud, and Drews}]{JJC}
Jeffrey Nichols, Jalal Mahmud, and Clemens Drews. 2012.
\newblock \href {https://doi.org/10.1145/2166966.2166999} {Summarizing sporting
  events using twitter}.
\newblock \emph{Proceedings of the 2012 ACM International Conference on
  Intelligent User Interfaces}, page 189–198.

\bibitem[{Nijholt et~al.(2003)Nijholt, Akker, and De~Jong}]{ANR}
Anton Nijholt, Rieks Op~Den Akker, and Franciska De~Jong. 2003.
\newblock \href
  {https://wwwhome.ewi.utwente.nl/~anijholt/artikelen/cuba2003.pdf} {Language
  interpretation and generation for football commentary}.
\newblock \emph{Actas del VIII Simposio Social}.

\bibitem[{Shaikh(2017)}]{classifiers}
Javed Shaikh. 2017.
\newblock \href
  {https://towardsdatascience.com/machine-learning-nlp-text-classification-using-scikit-learn-python-and-nltk-c52b92a7c73a}
  {Machine learning, nlp: Text classification using scikit-learn, python and
  nltk.}
\newblock \emph{Towards Data Science}.

\bibitem[{Zhang et~al.(2016)Zhang, Yao, and Wan}]{ZYX}
Jianmin Zhang, Jin-ge Yao, and Xiaojun Wan. 2016.
\newblock \href {https://wanxiaojun.github.io/acl16_sports.pdf} {Towards
  constructing sports news from live text commentary}.
\newblock \emph{AACL}.

\bibitem[{Zheng and Kudenko(2010)}]{MZDK}
Maliang Zheng and Daniel Kudenko. 2010.
\newblock \href {https://doi.org/10.4018/jgcms.2010100105} {Automated event
  recognition for football commentary generation}.
\newblock \emph{IJGCMS}, 2:67--84.

\end{thebibliography}
\bibliographystyle{acl_natbib}

\appendix

\section{Appendix}
Dictionary of the event categories and their corresponding labels :

\renewcommand{\arraystretch}{1}
\begin{table}[H]
\begin{center}
\begin{tabular}{ c c  }
 Labels & Type of event \\ \hline
 0 & No event \\ 
 1 & Attempt \\  
 2 & Corner \\ 
 3 & Foul \\ 
 4 & Yellow card \\ 
 5 & Second yellow card \\
 6 & Red card  \\ 
 7 & Substitution \\ 
 8 & Free kick won \\ 
 9 & Offside \\ 
 10 & Handball \\ 
 11 & Penalty conceded \\ 
 \end{tabular}
 \caption{\label{tab:table-name} Table of the different classes of our classification model and their labels.}
 \end{center}
 \end{table}

 \section{Appendix}
 You can find \href{https://github.com/yanismiraoui/Analyzing-sports-commentary-in-order-to-automatically-recognize-events-and-extract-insights}{here} the code that has been done during the elaboration of this study. 
 This code has been cleaned and annotated with comments explaining the methods and techniques used. Please read the \href{https://github.com/yanismiraoui/Analyzing-sports-commentary-in-order-to-automatically-recognize-events-and-extract-insights/blob/master/README.md}{README.md} file for more information.

  \section{Appendix}
  Moreover, you can also find \href{https://dash-models-commentary.herokuapp.com/}{here} a web application simulating some live commentary examples and their predictions based on the chosen model (SVM, XGBoost, ...). Please note: username is "admin" and password is "admin".
  The code of this web application is available \href{https://github.com/yanismiraoui/dash-models}{here} and has been deployed and hosted on a standard virtual machine.

\end{document}